\documentclass{article} 
\usepackage{iclr2023_conference,times}
\usepackage{graphicx}
\usepackage{subcaption}
\usepackage{algorithm}
\usepackage{algpseudocode}
\usepackage{amsmath}
\usepackage{amsfonts}
\usepackage{booktabs}
\iclrfinalcopy

\usepackage{xspace}
\usepackage[textsize=tiny,
]{todonotes}

\title{Estimating Residential Solar Potential using Aerial Data}

\author{Ross Goroshin, Alex Wilson, Andrew Lamb, Betty Peng, Brandon Ewonus, Cornelius Ratsch, \\
{\bf Jordan Raisher, Marisa Leung, Max Burq, Thomas Colthurst, William Rucklidge, Carl Elkin} \\
~~~~~~~~~~~~~~~~~~~~~~~~~~~~~~~~~~~~~~~~~~~~~~~~~~~~~~~~~~~~~~~~~~~~Google Inc \\
    \texttt{\{goroshin, alexwilson, andrewlamb, bettypeng, bewonus, cratsch} \\
    \texttt{ jraisher, marisaleung, mburq, thomaswc, wjr, celkin\}@google.com}
}

\begin{document}

\maketitle

\begin{abstract}
Project Sunroof estimates the solar potential of residential buildings using high quality aerial data. That is, it estimates the potential solar energy (and associated financial savings) that can be captured by buildings if solar panels were to be installed on their roofs. Unfortunately its coverage is limited by the lack of high resolution digital surface map (DSM) data. We present a deep learning approach that bridges this gap by enhancing widely available low-resolution data, thereby dramatically increasing the coverage of Sunroof. We also present some ongoing efforts to potentially improve accuracy even further by replacing certain algorithmic components of Sunroof's processing pipeline with deep learning. 
\end{abstract}

\section{Introduction}

\begin{figure}[h]
 \includegraphics[width=1.0\linewidth]{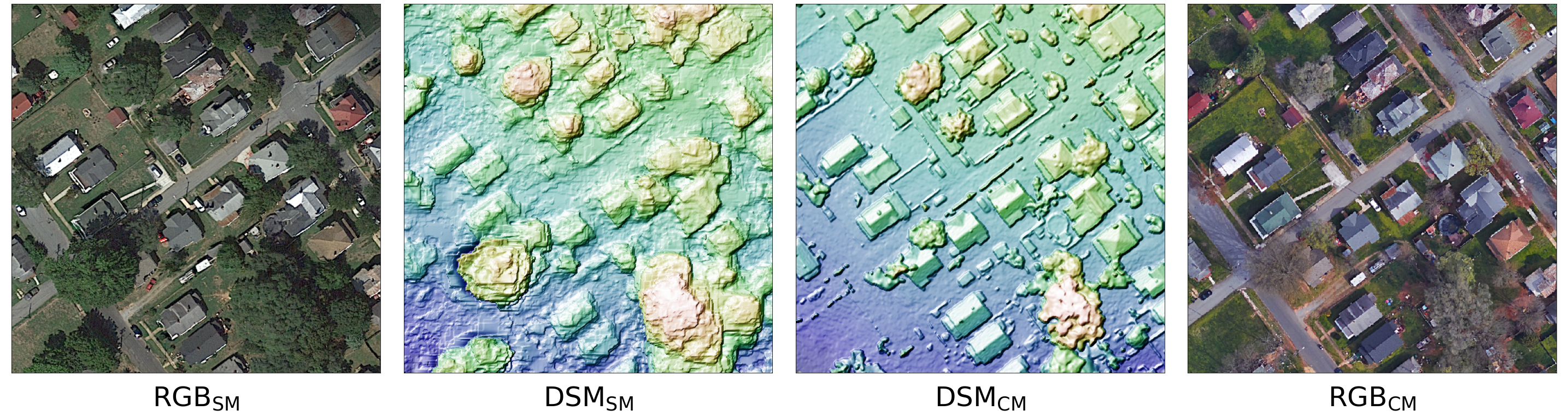}
 \caption{Low quality, sub-meter, imagery and its corresponding digital surface map are shown on the left (RGB$_\text{SM}$ and DSM$_\text{SM}$). High quality, centimeter-scale, of the same area collected at a different time, are shown on the right side (RGB$_\text{CM}$ and DSM$_\text{CM}$). The DSMs are rendered in 3D using the ``hillshade'' technique to better visualize geometric details.}
 \label{fig:data_overview}
\end{figure}

Sunroof enables potential customers and policymakers to make informed decisions regarding solar energy by providing accurate solar energy estimates for individual buildings. Since its release, we estimate that Sunroof has been used in over 1 million residential solar projects. Installers report that the use of Sunroof substantially increases the likelihood of solar adoption by their customers. 

In order to accurately estimate a building's solar energy potential it is necessary to first create a detailed 3D model of it and its surrounding area. Even tiny features like chimneys, air vents or AC units can impact the viable install area for solar panels. Of particular importance is the precise geometry of the roof, as the expected angle of incidence of sunlight has a major impact on annual generated energy. Sunroof mainly uses a digital surface model (DSM) and a digital terrain model (DTM) to simulate solar potential, which are currently obtained from aerial imagery \cite{Sole1996}. Both give elevations as functions of location on a grid, with the DTM giving the elevation of the terrain, and the DSM providing the elevation inclusive of all objects, such as trees and buildings. The DSM can be computed from overlapping images using standard stereo vision algorithms.

Sunroof relies on high quality, centimeter-level, aerial imagery in order to resolve details that are necessary for producing high accuracy solar estimates. However, even aerial imagery has varying degrees of quality. High flying aircraft equipped with a single camera can cover a large area at lower resolution, while lower flying aircraft equipped with calibrated multi-camera rigs are able to capture and register images at very high resolution. Other factors such as image registration quality and number of cameras used for stereoscopy, determine the signal to noise ratio of elevation data and therefore influence the effective resolution of the data. Figure \ref{fig:data_overview} shows example DSMs computed from low and high quality image inputs, which we will refer to as sub-meter (DSM$_{\text{SM}}$) and centimeter (DSM$_\text{CM}$) scale, respectively. Sunroof relies on DSM$_\text{CM}$ data to compute solar potential estimates. Unfortunately, high quality aerial imagery is much more limited in its coverage and update frequency. In Section \ref{sec:super-sunroof} we will discuss how this limitation was overcome and enabled Sunroof to use widely available lower quality data, thus expanding its coverage, and thereby potentially increase the rate of solar adoption in new areas. 

\section{The Sunroof Algorithm}
Sunroof estimates the solar power potential for the buildings in a given area by performing five major processing steps outlined in Algorithm \ref{alg:sunroof}. Steps (1-4) are involved in computing the viable solar potential of individual buildings. \emph{The resulting energy predicted by steps (1-3) was physically validated by the National Renewable Energy Laboratory (NREL) \cite{nrel}. Furthermore, the entire pipeline has been validated by several major solar install companies which have queried millions of addresses via our API.} Details of each step are described in Appendix \ref{sec:Sunroof}. 
\begin{algorithm}
\caption{Sunroof Algorithm}
\label{alg:sunroof}
\begin{algorithmic}
    \State $Inputs: \texttt{RGB, DSM, building footprints and probabilities}$
    \State $(1)~\texttt{footprints} \gets \texttt{SegmentBuildings(RGB, DSM, footprints, probabilities)}$
    \For{$\texttt{each building} \in \texttt{footprints}$}
        \State $(2)$ \texttt{Segment roof DSM into planes and remove obstacles}
        \State $(3)$ \texttt{Efficiently compute solar flux with fast ray-tracing} 
        \State $(4)$ \texttt{Compute panel layout \& power produced by each panel}
        \State $(5)$ \texttt{Financial calculations}
    \EndFor
    \State $Outputs: \texttt{Potential solar power and cost savings}$
\end{algorithmic}
\end{algorithm}

\section{Enabling Sunroof to Process Low Quality Data}
\label{sec:super-sunroof}
Our low quality data coverage is about $\sim 10 \times$ the area of the high quality data coverage, and is updated much more frequently. Therefore we seek a way to apply the Sunroof algorithm to the low quality data. As outlined in the previous section, Sunroof mainly relies on the DSM to compute the solar potential. Unfortunately, Sunroof does not generate accurate solar estimates from low quality DSMs.

In order to overcome this limitation we train a deep network to enhance the low quality DSM, by fusing information from multiple input modalities. If the enhanced DSM is a sufficiently accurate estimate of the high resolution DSM, it can be input to the unmodified Sunroof algorithm to estimate the solar potential. After presenting this approach and its results, we discuss a potential future improvement to the Sunroof algorithm itself, which replaces the graph-cut based roof segmentation (Subsection \ref{subsec:segmentation}) with a segmentation derived from an additional output head of our model. 

\subsection{Architecture}
The inputs to our enhancement model are: 1-a visible spectrum RGB image of the area, 2-a low quality DSM and DTM, 3-building probabilities (see Appendix \ref{subsec:footprints}). The outputs of our multi-head model are: 1-an enhanced DSM, 2-enhanced footprints, and 3-planar roof segments. All inputs, outputs, and targets have the same shape and spatial extent ($512 \times 512$ covering the same area) but inputs are generally lower quality (i.e. have lower underlying resolution and are more noisy than the corresponding high quality DSM). We use the corresponding high quality data (i.e. DSM, footprint, and roof segments) as the target for the corresponding output head. High and low quality data are often collected several years apart (``temporal skew''), therefore high quality targets should be regarded as imperfect/misaligned ground truth. For example, in Figure \ref{fig:data_overview} the tree at lower left of the DSM$_\text{SM}$ is missing in the DSM$_\text{CM}$. We use a UNet \cite{unet}-like architecture with a ResNet-50 encoder \cite{resnet} backbone pretrained on ImageNet classification \cite{imagenet}. Instead of using transposed-convolution, our architecture uses pixel-shuffle layers inspired by super-resolution architectures \cite{pixelshuffle}. The full architecture is depicted in Figure \ref{fig:arch}. 

\subsection{Enhanced DSM}
\begin{figure*}
 \includegraphics[width=1.0\linewidth]{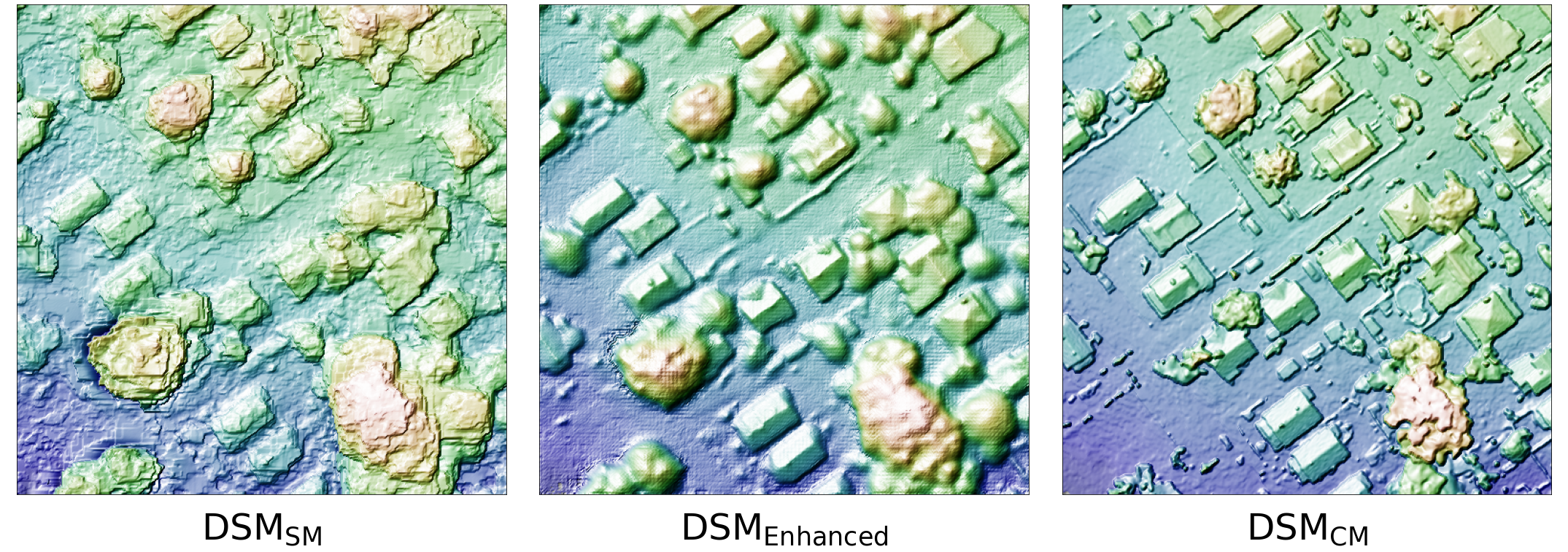}
 \caption{DSM enhancement by our model of the same area as Figure \ref{fig:data_overview}. DSM$_\text{SM}$ is input to the model while the DSM$_\text{CM}$ is used as ground truth.}
 \label{fig:dsm_enhance}
\end{figure*}
Inferring absolute elevation (or similarly depth) from monocular images is an ill-posed problem \cite{deigen}. Therefore, we train our network to infer relative elevation using a low resolution digital terrain map (DTM) as a reference \cite{dsm}. This is achieved by inputting DSM$_\text{SM}$ - DTM$_\text{SM}$ to our model, and then adding the DTM$_\text{SM}$ to the output before computing the error with the DSM$_\text{CM}$ as a target. Our loss consists of per-pixel modified $L_1$ regression and surface normal terms, specifically: 
\begin{align*}
L_1 &= |h_o - h_t - \lambda * \sum_x \sum_y \left[h_o - h_t\right]|\\
L_{sn} &= 1 - \frac{n_o \cdot  n_t}{||n_o||~||n_t||}
\end{align*}

The second term in the $L_1$ loss partially discounts the average deviation (over all spatial locations) between the output and target elevation maps. We have found that setting $\lambda=\frac{1}{2}$ yields the best performance in metrics discussed in Section \ref{sec:performance}. The terms $n_o$ and $n_t$ are the normal vectors computed at each point of the output and target DSMs, respectively. To compute the outward pointing normal vector we use the following expression:
\begin{equation*}
n = \left [-g_x, g_y, 1\right]
\end{equation*}
Where $g_x$ and $g_y$ denote the $x$ and $y$ components of the gradient of the DSM, respectively. These are computed using finite differences and local smoothing. Surface normal loss minimization has the effect of enhancing high frequency details of the output DSM, as well directly enforcing that the normal vectors of roof segments derived from the output DSM are accurate. Similar approaches have been used in \cite{dsm} and \cite{im2elevation}. 

\subsection{Additional Outputs}
Our network also outputs refined building probability and semantic roof segmentation maps. Building probabilities corresponding to low and high quality images are obtained automatically using another network specifically trained to perform building segmentation (similar to \cite{mmeka}). Our network is trained to enhance building probabilities corresponding to low quality data to match those corresponding to high quality data using a binary cross-entropy loss. Roof plane segmentation is achieved using a single shot affinity mask approach with graph partition post-processing to obtain instance segments similar to \citep{segmentation}. Briefly, the $N$-dimensional output at each location predicts whether its $N$ neighboring pixels belongs to the same instance label. Instance segments are obtained by applying a graph partitioning algorithm, to a graph whose edge weights are computed from the affinity maps output by the network (see Figure \ref{fig:seg_output}). Ground truth roof instance segments are obtained by applying Sunroof's graph-cut algorithm to the corresponding high quality DSMs described in Appendix \ref{subsec:segmentation}. Results presented in the next section were obtained by feeding enhanced DSMs and building probabilities to the Sunroof algorithm.  Note that despite being trained to output the roof segmentation, the results presented in the next section used Sunroof's original segmentation approach described in Appendix \ref{subsec:segmentation}. 

\section{Performance and Future Work}

\begin{table}
\renewcommand{\arraystretch}{1.2}
    \begin{center}
        \begin{tabular}{|c c c c|} 
        \hline
         Data & MAPE & MAPE@5kW & Temporal Separation \\ 
         \hline
         USA & $29.14 \% \pm 10.50$  & $4.62 \% \pm 1.58$ & $<1$ year  \\ 
         \hline
         EU & $31.00 \% \pm 9.36$ & $6.43 \% \pm 1.89$ & $<1$ year \\
         \hline
         Temporal & $20.10 \% \pm 12.71$ & $4.43 \% \pm 2.58$ & 1-5 years \\
         \hline
        \end{tabular}
    \end{center}
    \caption{MAPE and MAPE@5kW errors in the EU and USA. The Temporal row shows the error due to temporal skew.}
\label{table:results}
\end{table}
\label{sec:performance}
We use two, physically grounded, performance measures to evaluate the performance of the refinement model (see Table \ref{table:results}). The first measures the percentage mean absolute power error (MAPE), that is the error between the total power, over the course of one year, predicted by Sunroof corresponding to the refined data and high quality data over the same area. Total power implies tiling the entire viable roof area in solar panels (see Figure \ref{fig:solar_panels}), something typical consumers rarely do. Therefore we introduce a second error measure, MAPE@5kW which measures the error corresponding to a much smaller, but more typical 5kW array ($\approx 10$ panels). These panels are optimally positioned by Sunroof in the most productive areas of the roof predicted using the solar flux (Figure \ref{fig:flux}). Thus the MAPE effectively measures the error for the entire roof, while the MAPE@5kW only measures the error corresponding to the most productive portions of the roof. Our enhancement model was trained on data collected over cities in the Southeastern USA, excluding any cities in Florida, which we reserved for validation. Our dataset is limited to cities where high/low quality pairs are available. The results presented in Table \ref{table:results} are over test cities which are not present in the training or validation sets. We also report performance on Western European cities. To minimize temporal skew between inputs and ground truth (e.g. new construction, seasonal foliage, etc.), we selected high/low quality dataset pairs that were collected at most one year apart. Finally, in order to estimate the effect of temporal skew we evaluated the error between \emph{high quality assets} collected between 1-5 years apart. This evaluation does not involve our enhancement model but simply compares Sunroof's predictions corresponding to temporally separated high quality datasets. This evaluation leads to a substantially lower MAPE value, which confirms that despite some temporal skew, the enhanced data still lags real high quality data in performance by about 10\%.  In ongoing work, we hope to achieve even better performance by replacing it with our model's output. We have found that the segmentation is extremely sensitive to the enhanced DSM and is thus a major source of MAPE error. 

\appendix

\section{Sunroof Details}
\label{sec:Sunroof}


\begin{figure}[ht]
\begin{subfigure}{.5\textwidth}
  \centering
  \includegraphics[width=0.95\linewidth]{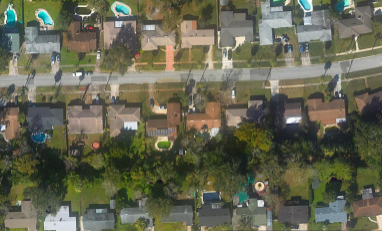}
  \caption{RGB}
  \label{fig:rgb}
\end{subfigure}%
\begin{subfigure}{.5\textwidth}
  \centering
  \includegraphics[width=0.95\linewidth]{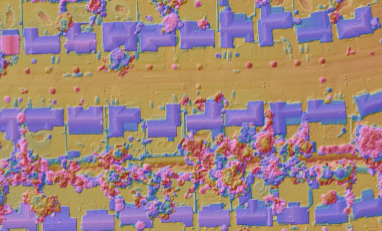}
  \caption{DSM}
  \label{fig:dsm}
\end{subfigure}
\begin{subfigure}{.5\textwidth}
  \centering
  \includegraphics[width=0.95\linewidth]{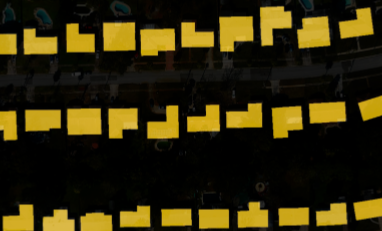}
  \caption{Building Footprints}
  \label{fig:footprints}
\end{subfigure}
\begin{subfigure}{.5\textwidth}
  \centering
  \includegraphics[width=0.95\linewidth]{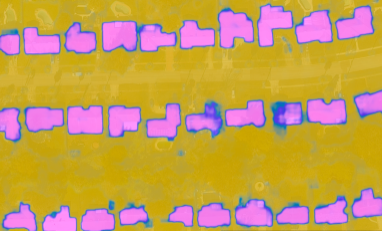}
  \caption{Building Probabilities}
  \label{fig:probs}
\end{subfigure}
\caption{Illustrations of the inputs to Sunroof.}
\label{fig:fig}
\end{figure}

\begin{figure}[ht]
\begin{subfigure}{.5\textwidth}
  \centering
  \includegraphics[width=0.95\linewidth]{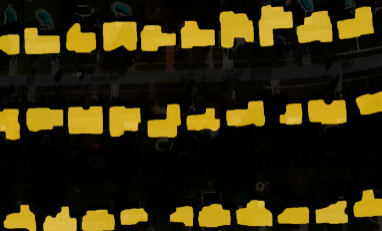}
  \caption{Improved Building Footprints}
  \label{fig:improved}
\end{subfigure}
\begin{subfigure}{.5\textwidth}
  \centering
  \includegraphics[width=0.95\linewidth]{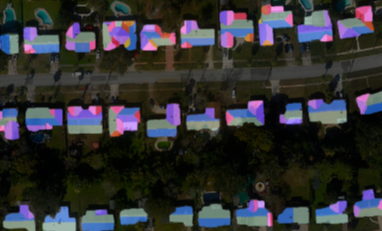}
  \caption{Roof Segments}
  \label{fig:segments}
\end{subfigure}
\begin{subfigure}{.5\textwidth}
  \centering
  \includegraphics[width=0.95\linewidth]{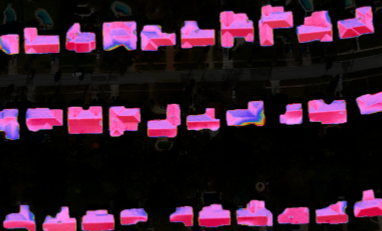}
  \caption{Solar Flux}
  \label{fig:flux}
\end{subfigure}
\begin{subfigure}{.5\textwidth}
  \centering
  \includegraphics[width=0.95\linewidth]{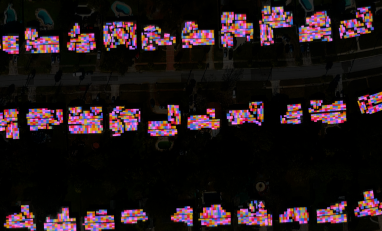}
  \caption{Solar Panel Layout}
  \label{fig:solar_panels}
\end{subfigure}
\caption{Illustrations of the outputs of the processing steps of Sunroof.}
\label{fig:fig}
\end{figure}

\subsection{Building Footprint Refinement}
\label{subsec:footprints}
An initial set of footprints and probabilities are input to the Sunroof algorithm. Both of these are output by other, separately trained, models. Building footprints are rough polygons, but often separate individual residential addresses although they may appear connected. Building probabilities take on values close to 1.0 where the corresponding pixel is likely to belong to a building. A graph is created whose edge weights are computed by fusing information from the footprints/ probabilities, and DSM (height discontinuities captured in the DSM are often good cues to detect the presence of buildings). Footprint refinement is performed by running a graph-cut algorithm on this graph. The refined footprints often remove ``tree overhangs'' or buildings that are entirely occluded by trees, thanks to information from the building probabilities, while preserving separation between residences using information from the footprints. 

\subsection{Roof Segmentation and Obstacle Removal}
\label{subsec:segmentation}
Next, the roof pixels in the DSM are fitted to a small number of planes using a RANSAC \cite{cantzler1981random} algorithm.  This is essential as solar panels can only be laid out on flat roof segments.  After the RANSAC procedure, the points assigned to each segment are refined using a graph cut \cite{greig1989exact} approach -- if a point could reasonably be assigned to multiple planes, the graph cut will prefer to assign it to the same plane as its neighbors.  Specifically, the graph cut algorithm attempts to minimize a cost function consisting of: (i) the projection distance from a point to its assigned plane, (ii) a second cost that minimizes the number of planes with similar normal vectors. This is intended to make it harder for two very similar planes to partition a flat area of the roof. The cost can be expressed as:

\begin{equation}
    \sum_p \sum_{{\bf P}} \left[ d(p, {\bf P}_p) + m \sum_{\substack{q \in \mbox{{\bf N}(p)}}} 1 + max \left( 0, \hat{n}_{{\bf P}_p} \cdot \hat{n}_{{\bf P}_q} \right) \right]
\end{equation}

Where $d(p, \bf{P}_p)$ denotes the projection distance when point $p$ is assigned to plane ${\bf P}$, ${\bf N}(p)$ denote the set of points neighboring point $p$, and $\hat{n}_{{\bf P}_p}$ denotes the normal vector to plane $\bf P$. After running the graph cut, each roof plane is refit based on its new points, and the graph cut plus refitting procedure is repeated several times. Finally, the roof segments are filtered by size to remove tiny segments. This step also removes roof obstacles such as air-vents and air-conditioners. 

\subsection{Solar Flux}
The solar flux calculation estimates the incident solar energy on a building over the course of a year (irradiance). Factors that affect solar flux include: the latitude, pitch angle of the roof segments, and surroundings which may occlude sunlight and cast shadows on the roof (usually trees or other buildings). The flux calculation is parallelized by partitioning the data into tiles. These tiles overlap, with each tile having a core plus margins. The margins are used so that nearby obstructions are taken into account when calculations are performed on the core. This means that the effects of distant occluders outside the margin area, such as distant mountain ranges, will not be factored into the calculation. The main flux calculation is performed using a method similar to \cite{timonen2010scalable}, and its computational complexity is linear in the number of pixels, with a constant that depends on latitude, compared to $O(n^3)$, for direct ray tracing. The irradiance is summarized as two quantities: Direct Normal Irradiance (DNI), which is sunlight received directly from the sun, and Diffuse Horizontal Irradiance (DHI) which is indirect sunlight received by a flat plane. Both are measured in units of $\mbox{\textit{Watts}}/m^2$. DNI and DHI are obtained from publicly available datasets, such as the National Solar Radiation Database (NSRDB) \cite{nsrdb}. Finally, both of these are further attenuated using a correction factor derived from the air temperature and wind speed using the model from \cite{schwingshackl2013wind}. This reflects the decrease in silicon solar panel efficiency in elevated temperatures. 

\subsection{Optimal Panel Layout and Power Prediction}
\begin{figure}[!h]
 \centering
 \includegraphics[width=0.75\linewidth]{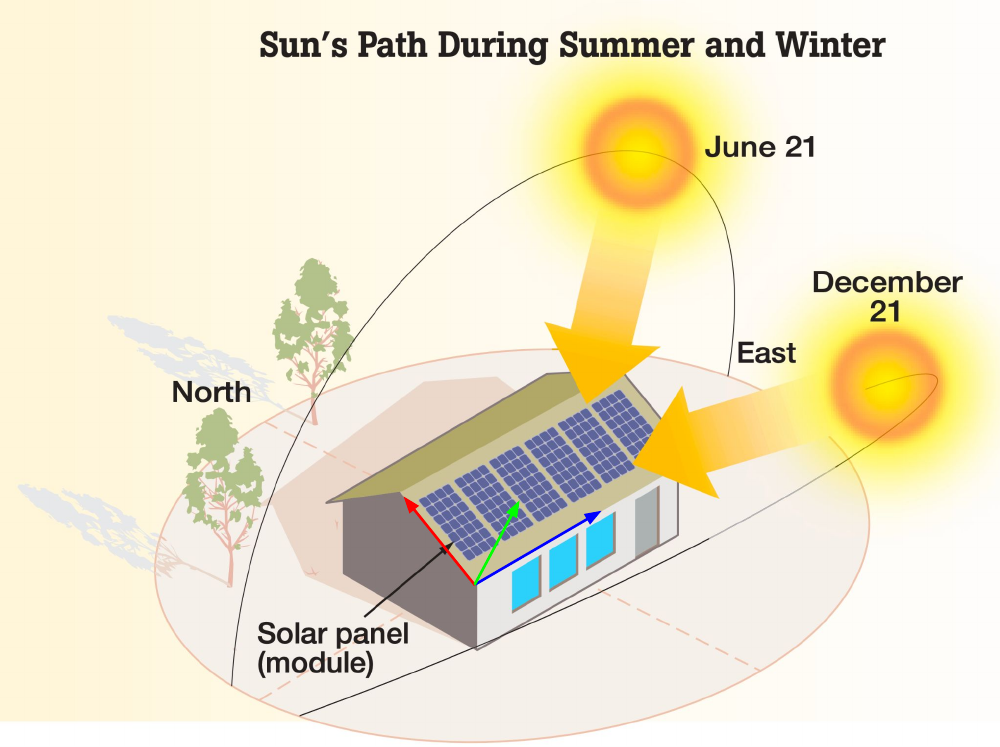}
 \caption{Depiction of irradiance and roof coordinate system used for panel layout. Red, green, and blue vectors are the up-slope, normal, and eaves vectors, respectively. Image credit: U.S. Department of Energy. }
 \label{fig:house}
\end{figure}

In order to get an upper bound on the solar potential of buildings, the solar panel placement algorithm tiles a roof with as many panels as the viable roof area can support. Viable roof areas include roof segments that are not overly steep and are free of obstacles. For sloped roof segments, solar panels are laid out in the 2D coordinates defined by the ``eaves'' and ``up-slope'' vectors (see Figure \ref{fig:house}). If the unit normal is $ \hat n = \left[n_x, n_y, n_z\right]$ then the eaves and up-slope vectors are defined as:

\begin{align}
    \hat e =& \left[\frac{n_y}{\sqrt{n_x^2+n_y^2}}, -\frac{n_x}{\sqrt{n_x^2+n_y^2}} ,0 \right] \nonumber \\
    \hat u =& ~\hat{e} \times \hat{n} \nonumber
\end{align}
For a horizontal roof ($n_x=n_y=0$), the eaves and up-slope vectors are chosen arbitrarily.

Panels are tiled over these flat roof segments in a way that maximize both energy production and compactness of the layout -- roughly minimize the area of the rectangle that bounds all panels in a given roof segment. The flux allows simple estimation of the expected power generated by each panel, this facilitates finding optimal configurations for smaller solar installations. For example, Sunroof can be used to find the optimal panel layout of a, more typical, 5kW array consisting of 10-20 panels. 

\section{Additional Figures}

\begin{figure}[!ht]
\begin{subfigure}{.5\textwidth}
  \centering
  \includegraphics[scale=0.25]{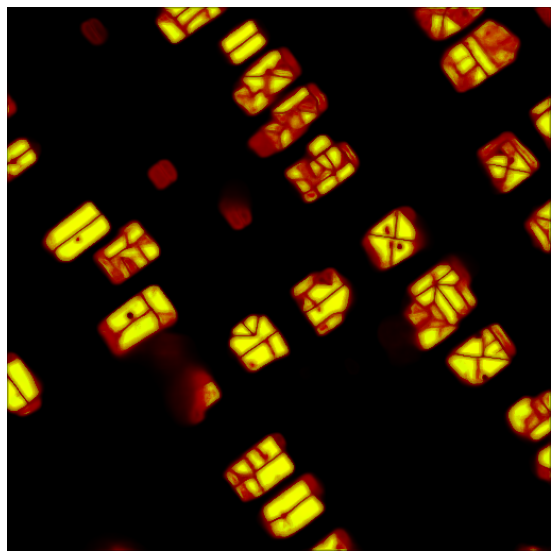}
  \caption{One Channel from the Affinity Mask Output}
  \label{fig:rgb}
\end{subfigure}
\hfill
\begin{subfigure}{.5\textwidth}
  \centering
  \includegraphics[scale=0.25]{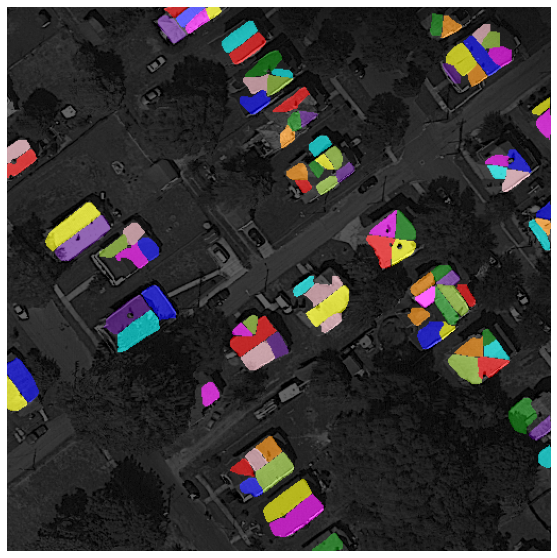}
  \caption{Instance Segments}
  \label{fig:dsm}
\end{subfigure}
\caption{Instance roof segments are obtained by post-processing the affinity mask output.}
\label{fig:seg_output}
\end{figure}

\begin{figure}[!htb]
 \includegraphics[width=1.0\linewidth]{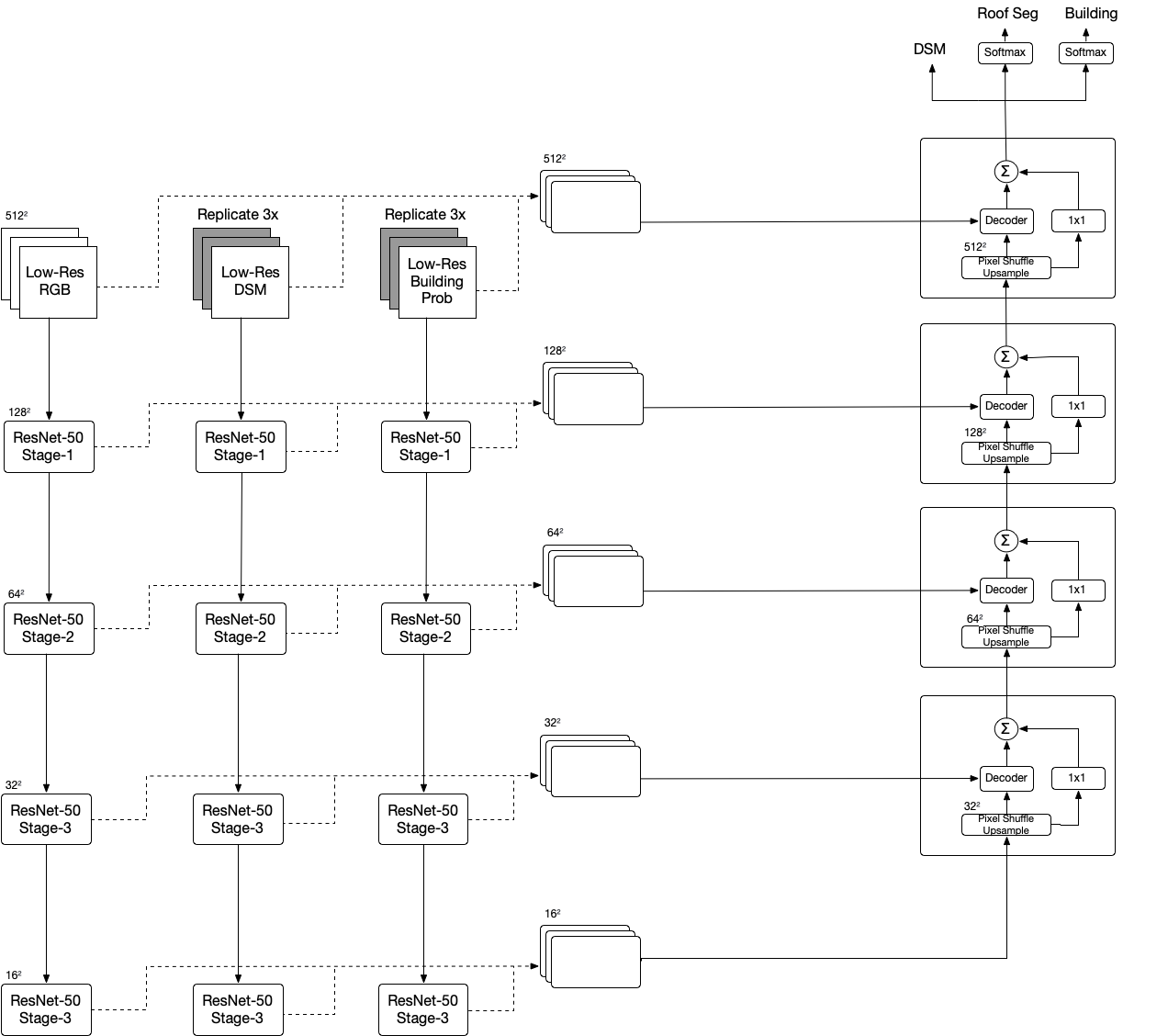}
 \caption{Sunroof UNet architecture}
 \label{fig:arch}
\end{figure}
\clearpage
\bibliographystyle{plainnat}
\bibliography{references}

\end{document}